\documentclass{article}
\usepackage{spconf,amsmath,graphicx,color,booktabs}
\usepackage[table, dvipsnames]{xcolor}

\usepackage{makeidx} 
\usepackage{changes}	
\usepackage{todonotes} 

\definechangesauthor[color=Red, name=\textbf{Alejandro}]{AC}

\usepackage{todonotes}
\makeatletter
\setremarkmarkup{\todo[color=Changes@Color#1!20,size=\scriptsize]{#1: #2}}
\makeatother

\title{All the people around me: face discovery in egocentric photo-streams}

\name{Maedeh Aghaei, Mariella Dimiccoli, Petia Radeva}
\address{University of Barcelona, Computer Vision Center}

\begin{document}

\maketitle

\begin{abstract}

\end{abstract}

Given an unconstrained stream of images captured by a wearable photo-camera (2fpm), we propose an unsupervised bottom-up approach for automatic clustering appearing faces into the individual identities present in these data. The problem is challenging since images are acquired under real world conditions; hence the visible appearance of the people in the images undergoes intensive variations. Our proposed pipeline consists of first arranging the photo-stream into events, later, localizing the appearance of multiple people in them, and finally, grouping various appearances of the same person across different events. Experimental results performed on a dataset acquired by wearing a photo-camera during one month, demonstrate the effectiveness of the proposed approach for the considered purpose.


\section{Introduction}
\label{sec:intro}

Face discovery, also known as face clustering, is the task of grouping face images in a dataset into either known or unknown number of disjoint groups. Face discovery is a suitable task for cases where the identity of people in the dataset is not available. Its applications range from the interactive photo album tagging \cite{lee2011face,zhu2011rank,xia2014face} to different aspects in social media \cite{wang2012unified}. 

Recently, the inclination of  people towards the use of wearable cameras to automatically record their moments has considerably increased \cite{bolanos2016toward}. Among the available wearable cameras, photo-cameras that take pictures at a lower frame-rate (i.e. 2fpm), without needing to recharge for several consecutive days, are more suitable for long time acquisition. Images collected over a long period of time contain valuable information about the lifestyle of the user. From a memorability perspective \cite{holland2010emotion}, the most prominent pictures among a large amount of captured images are those to which the user attaches specific emotions, which indisputably involve social interactions \cite{lopes2004emotional,khosla2012memorability}. 
\begin{figure}[htb]
\centering
\centerline{\includegraphics[width=8.5cm]{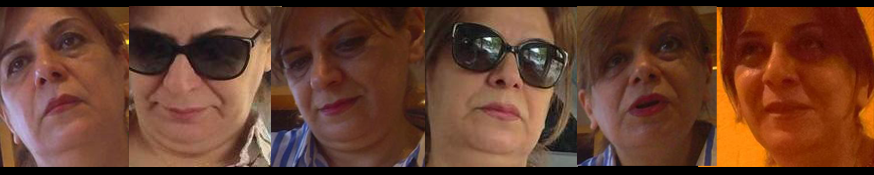}}
\caption{Example of randomly selected images of a same person in different events in our dataset.}
\label{fig:example}
\end{figure}
Due to the psychological effect of social interactions, the medical applications of face discovery in egocentric photo-streams are inevitable \cite{dhand2016accuracy,hodges2011sensecam,berry2007use}. Face discovery is also useful to unveil less noticed matters about the social life of the user: with whom does he/she interacts the most? how many times he/she has seen his/her friends last month?

From the perspective of computer vision, the most challenging problem in face discovery is how to find correctly the alternative occurrences of the same face in a dataset. In fact, appearing faces inevitably undergo intensive changes when the camera is worn under free living conditions, due to variations in lighting, pose, viewpoint, facial accessories, and so forth (see Fig \ref{fig:example}). In this paper, we proposed a fully unsupervised approach for face discovery from egocentric images collected by a wearable photo-camera over a long period of time. In this context, face is the most discriminating feature of a person since, depending mostly on the clothing, a person appearance may change drastically in different days or even at different times of the day. To cope with the extreme intra-class variability of faces, we propose to first track the appearance of multiple faces into a same event using \cite{aghaei2016multi}, and then considering both the \textit{inner-track} and \textit{inter-track} constraints, to cluster similar faces across the events into unknown number of groups. 

The rest of the paper is organized as follows: in the next section we review the state-of-the-art methods on face clustering; in section \ref{sec:methododlogy} we detail our proposed approach for face clustering in egocentric photo-streams, in section \ref{sec:exp} we introduce the dataset used in this paper as well as the experimental setting and we discuss the experimental results. Finally in section \ref{sec:concl}, we summarize the content and the contributions of our work.

\section{Related work}
\label{subsec:SoA}

Face clustering is a largely unconstrained problem and rich body of work in the literature has focused in finding how to exploit characteristics of the dataset or of the particular application to constrain it. The most common applications are interactive tagging of photo albums \cite{lee2011face,zhu2011rank,xia2014face} and video organization \cite{xiao2014weighted,zhang2016joint}. In the context of face discovery in photo albums, Lee et al. \cite{lee2011face} introduced a new constraint known as \textit{social context of co-occurring people}, following which people of the same social context often appear together. For example, faces of the family members usually tend to co-occur even in different photos. The system first trains a separate detector for each individual and later, uses the detector to discover novel face clusters by taking advantage of co-occurrence constraints. In the same scenario, Zhu et al. \cite{zhu2011rank} presented a Rank-Order distance to measure the dissimilarity between two faces. This work exploits the fact that faces of the same person usually form close sub-clusters in the feature space. A similar idea is proposed by Xia et al. \cite{xia2014face}, who exploited two constraints: an individual only may appear once in a picture, and the number of instances of a same person must be lower than the total number of pictures. The problem is then formulated as a constrained K-Means, which is solved through Minimum Cost Flow linear network optimization strategy. Imposing constraints to achieve more accurate clustering is observed in several other works attempting to cluster faces in videos. Xiao et al. \cite{xiao2014weighted} proposed a Weighted Block-Sparse Low Rank Representation (WBSLRR) which learns a low rank data representation, while considering two defined prior constraints. First, the inner-track constraint states that any two faces in the same face track belong to the same person. Therefore clustering is first performed over face-tracks instead of individual faces. Second, the inter-track constraint that states face-tracks belonging to faces that appear in the same frame, does not belong to the same person. A similar idea has been employed by Cinbis et al. \cite{cinbis2011unsupervised}, to learn a distance metric for face identification in videos that pulls close together faces in an inner-track relation, and pushes away those in inter-track relation. More recently, as in many other computer vision tasks, deep features proved their efficiency in data representation for face clustering \cite{schroff2015facenet,zhang2016joint}. However, deep learning based approaches are supervised and hence require a previous learning stage involving identity-labeled faces. Therefore, they are most suited for face re-identification. 

\section{METHODOLOGY}
\label{sec:methododlogy}

Given a large unconstrained photo-stream captured by a wearable camera, we propose a face clustering approach by leveraging inner-class and inter-class constraints derived from the face tracking of people across the photo-stream.

\begin{figure}[htb]
\centering
\centerline{\includegraphics[width=8cm,height=3.8cm]{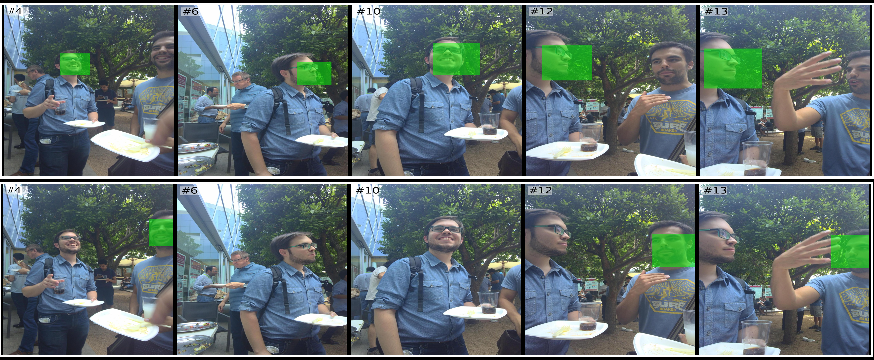}}
\caption{Each row is the resulting prototype of tracking by eBoT \cite{aghaei2016multi} over a sequence of two people.}
\label{fig:ebot}
\end{figure}
\subsection{Multi-face tracking in egocentric photo-streams}

To overcome the challenges imposed by the free motion of the camera and by its low temporal resolution, we previously proposed a multi-face tracking method \cite{aghaei2016multi}. Prior to any computation, first, a temporal segmentation algorithm \cite{dimiccoli2016sr} is applied to extract segments characterized by similar visual properties. Later on, a face detector is applied on all the frames of a segment to detect visible faces on them \cite{amos2016openface}. Based on the ratio between the number of frames with detected faces and the total number of frames of the segment, we extract segments containing trackable persons. The next steps are applied on these extracted segments, hereafter referred to as \emph{sequences}. Our multi-face tracker generates a tracklet for each detected face by finding its correspondences along the whole sequence, and then takes advantage of the tracklet redundancies to deal with the unreliable ones. Similar tracklets are grouped into the so-called extended-bag-of-tracklets (eBoT), which is aimed to correspond to a specific person. Finally, a prototype tracklet is extracted for each eBoT, and occlusions are estimated by relying on a measure of confidence. A final prototype keeps the bounding boxes of face occurrences of one individual along that sequence in the photo-stream, so in the case that two persons appear in a sequence, eBoT outputs two separate prototypes that localize face occurrences of each individual, separately (see fig. \ref{fig:ebot}). Due to the characteristics of the camera, faces appear in variety of views and in different ambient conditions (see fig. \ref{fig:example}). We treat all the observed occurring variations of the same face in a sequence as a unique representation of the same face for face discovery in the whole dataset. Hereafter, we refer to each bounding box in a sequence as a \textit{face-example} and define all the bounding boxes as the output of a sequence as a \textit{face-set}.

\subsection{Face discovery in egocentric photo-streams}
\label{subsec:FaceDiscovery}

Unlike the majority of face discovery frameworks that solely rely on pair-wise comparison of face-examples at time to find face matches, our system is built upon a tracking framework that provides us with a set of correct variations of the same face in one sequence. In this way, the proposed algorithm reshapes the face discovery challenge from face-pair comparison, to face-set-pair comparison. In our  approach, the deterministic factor in deciding whether two different face-sets belong to the same person is defined through a measure of dissimilarity. We first calculate the dissimilarity between all the possible pairs of face-sets, and then, based on these measurements, we employ a hierarchical clustering technique to discover the most similar face-sets.

\textbf{Dissimilarity between two face-sets:} 
For simplicity, let us suppose that given two face-sets, say $R$ and $T$, we want to measure the dissimilarity between \textit{target}, $T$, and  the \textit{reference}, $R$. Let $l(R)$ and $l(T)$ be the lengths of $R$ and $T$, respectively. Let $r_i \in R$ be the $i-$th face example in the $R$, where $i = 1,\ldots,l(R)$ and $t_j \in T$ be the $j-$th face example in the $T$, where $j = 1,\ldots,l(T)$.
To compute the dissimilarity between $R$ and $T$, we first define two similarity matrices: $S^R$ representing the similarity between all possible pairs of face-examples in $R$, and $S^T$ representing the similarity between face-examples in $R$ and face-examples in $T$. We compute $S^R$ as to build a baseline about how similar faces inside a face-set are. The similarity between two face-examples is measured by their average deep-matching score \cite{weinzaepfel2013deepflow}. The deep-matching is a descriptor matching algorithm, built upon a multi-stage architecture with interleaving convolutions and max-pooling layers and uses dense sampling to retrieve correspondences with deformable patches. More specifically, instead of using SIFT patches as descriptors, each SIFT patch is split into four quadrants and, assuming independent motion of each of the four quadrants, the similarity is computed to optimize the quadrant positions of the target descriptor. As a consequence, the descriptor is able to deal with various kinds of image deformations, including scaling factors and rotations. Denoting by ${\Delta (x,y)}$ the value of the deep-matching between $x$ and $y$, the elements of $S^R$ are defined as $s^R_{i,k} = \Delta(r_i,r_k) , i,k=1, \ldots, l(R)$ and the elements of $S^T$ as $s^T_{i,m} = \Delta(r_i,t_m)$, with $i =1, \ldots, l(R) , m= 1, \ldots ,l(T)$. Finally, the dissimilarity $\delta(R,T)$ between $R$ and $T$ is calculated as the absolute difference between the median value of $S^R$ and $S^T$, say $\varphi ^R$ and $\varphi ^T$, respectively:
\begin{equation}\label{eq:diss}
 \delta(R,T) = \left | \varphi ^R - \varphi ^T \right |
\end{equation}

\textbf{Clustering of face-sets:} To cluster face-sets based on their dissimilarity, we used agglomerative clustering, a hierarchical bottom-up approach that repeatedly merges pairs of clusters based on a measure of dissimilarity to form larger clusters. In this work, the initial clusters are face-sets and Eq. (\ref{eq:diss}) is used to measure the dissimilarity between face-set-pairs. All dissimilarity relations between face-set-pairs are encoded by the matrix $\mathcal{D} \in \mathcal{R}^{N \times N}$, where $N$ is the total number of face-sets. To take into account the fact that face-sets extracted from the same sequence should belong to different subjects, we force the dissimilarity between these face-sets to be maximal.  Specifically, we introduce the constraint matrix $\mathcal{C}\in \mathcal{R}^{N \times N}$, where its elements $c_{m,n} = 1$ if the face-sets $r_m \in R$ and $t_n \in T$ were extracted from the same sequence and $c_{m,n} = 0$, otherwise. We then multiply each element of $\mathcal{D}$, say $d_{m,n}$, by the weight $w_{m,n} = c_{m,n} + 1$.

\begin{figure}[htb]
\centering
\centerline{\includegraphics[width=7cm]{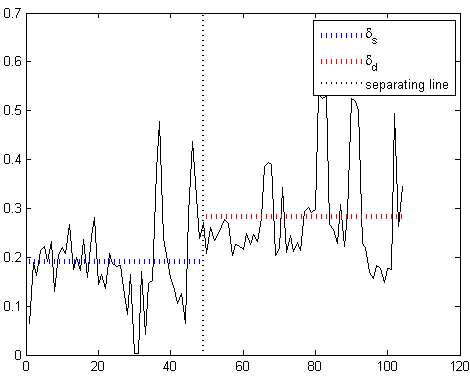}}
\caption{Threshold estimation on a training set: left side of the separating line shows $\delta_s (R,T)$ values, and right side of it shows $\delta_d (R,T)$ values. The horizontal lines are the median of $\delta(R,T)$ values in each section.}
\label{fig:threshold}
\end{figure}

To determine the cut-off threshold, that is, when to stop merging clusters at a selected precision, we  measured $\delta (R,T)$  of various face-sets in two manner: first, $\delta_s (R,T)$, where $R$ and $T$ are different face-sets of the same person, and second, $\delta_d (R,T)$, where  $R$ and $T$ belong to two different people. The cut-off threshold $\theta$ is taken as the median value of all the values of $\delta_s (R,T)$. These calculations are performed over a training dataset consisting of 100 face-sets. Fig. \ref{fig:threshold} shows the $\delta_s (R,T)$ values on the left and $\delta_d (R,T)$, values on the right, over the training dataset, where the vertical is the separating line between them and the horizontal lines are the median of $\delta(R,T)$ values in each section.

\section{Experimental results}
\label{sec:exp}
\subsection{Dataset}
\label{ssec:data}

The dataset used for validating our proposed model consists of 2,639 images captured by one user wearing the Narrative Clip camera during 30 days. It contains 2,033 faces belonging to 40 persons, whose bounding boxes have been manually annotated. In average, each person appears in 6 sequences and 3 days. There are 35 sequences with more than one person appearing on them over 113, in total. A separate dataset is used to select a cutoff value discussed in the previous subsection. It is acquired by 8 users; each user wore the Narrative clip for a number of non-consecutive days over a total of 100 days period, collecting 30,000 images, where 3,000 images of them contain a total number of 100 different trackable persons. Sequences in both datasets have different lengths, varying from 10 to 40 frames and they have been acquired in real world conditions, including inside and outside scenes.

\begin{table*}[]
\centering

\caption{Percentage of NMI and ARI values}
\label{table1}
\begin{tabular}{@{}lllllllll@{}}
\toprule
 & \textbf{M1} & \textbf{M2} & \textbf{M3} & \textbf{M4} & \textbf{M5} & \textbf{M6} & \textbf{M7} \\ \midrule
\multicolumn{1}{c|}{\textbf{NMI}} & \multicolumn{1}{l|}{24.31} & \multicolumn{1}{l|}{21.18} & \multicolumn{1}{l|}{58.35} & \multicolumn{1}{l|}{68.95} & \multicolumn{1}{l|}{19.21} & \multicolumn{1}{l|}{78.79} & \multicolumn{1}{l|}{\textbf{83.68}} 
\\ \cmidrule(l){2-8} 
\multicolumn{1}{c|}{\textbf{ARI}} & \multicolumn{1}{l|}{00.59} &  \multicolumn{1}{l|}{00.21} & \multicolumn{1}{l|}{31.66} & \multicolumn{1}{l|}{01.49} & \multicolumn{1}{l|}{00.42} & \multicolumn{1}{l|}{23.44} & \multicolumn{1}{l|}{\textbf{33.84}}
\\ \bottomrule
\end{tabular}
\end{table*}


\subsection{Baseline}

In this section, we evaluate state-of-the-art methods with different settings over our dataset. The following is a brief description of each setting. 

\textbf{M1 (WBSLRR):} the proposed method in \cite{xiao2014weighted} applied on the face-sets obtained by applying eBoT.


\textbf{M2 (Spectral, Open-face, Face-pairs):} an implementation of face recognition with deep neural networks based on the work proposed in \cite{schroff2015facenet}, known as OpenFace \cite{amos2016openface}. Faces are first detected using a pre-trained model for face detection. Second, they are transformed in an attempt to make the eyes and bottom lip appear approximately in the same location on each image. Third, a deep neural network is used to embed the face on a 128-dimensional unit hypersphere. Finally, spectral clustering method is used to group faces into groups corresponding to different subjects. 

\textbf{M3 (Agglomerative, Open-face+Euclidean, Face-pairs):} same setting as M2 is employed, despite variations in the forth step. In this setting, Agglomerative clustering is applied over the pair-to-pair Euclidean distance between 128-dimensional face features.

\textbf{M4 (Spectral, Open-face, Face-set-pairs):} to try to improve the results, we used as initial clusters the face-sets resulting from applying eBoT. A unique 128-dimensional feature vector as the mean value of all the 128-dimensional faces feature vectors is representing each face-set.

\textbf{M5 (Agglomerative, Open-face+Euclidean, Face-set-pairs):} similar setting to M4 for face-set representation is employed. Agglomerative clustering is then employed to cluster face-sets based on their distance from each other.

\subsection{Evaluation measurements}
\label{ssec:metrics}
To compare our proposed method with the baseline, we used two distinct widely known measurement techniques for clustering evaluation with known true-labels. The first metric, we used, is Normalized Mutual Information (NMI), that measures the mutual information between the labels predicted by the classifier and the true-labels, ignoring permutations. The second metric is Adjusted Rand Index (ARI), that measures how similar the labels predicted by the classifier are to the true-labels. Mathematically, ARI is related to the accuracy. It evaluates on a pairwise-basis if two sets of labels are incorrectly grouped so its value is representative of the true clustering result. Both measurements have values ranging from $0$ to $1$, with $1$ indicating that the clustering result perfectly matches the ground truth.

\subsection{Discussion}

Although, NMI and ARI validate the results in distinct ways, both follow the same trend as it can be observed for different methods in Table \ref{table1}. M1 to M5 are the baselines introduced previously, M6 is the proposed model without considering the inter-track constraints, and M7 is the complete pipeline for the proposed method, as described in Sec \ref{subsec:FaceDiscovery}. As expected, constraining the problem by exploiting inter-track constraints (M7), allows to improve the accuracy up to 10$\%$ considering ARI with respect to the same approach without inter-track constraints (M6).
Open-face is a robust method for extraction of face features. However, as it can be observed, the proposed method employing the deep-matching approach can grasp a more robust idea of the similarity measure between face-example pairs which is proved by higher NMI and ARI values comparing M2 to M5 with M6 and M7. Additionally, WEBSLRR, despite its robust pipeline for face clustering in controlled environments, performs poorly on our dataset. We consider that this is due to using only pixel intensities for the image representation. The experiments performed in this work have unveiled how challenging it is to cluster faces appearing in photo-streams captured by a wearable camera under free-living conditions. Indeed, face appearances may change even along the same event since people take out or put on their accessories such as glasses, a hat, make up and so forth, making the problem very challenging. However, from our experiments, we can say that the unbalanced number of images per individual in a face-set is the most challenging problem faced by face clustering. In this work, we aimed to study only facial attributes, disregarding any additional information. This analysis is important when the additional features are not either available, because of the nature of the applications or they are costly to provide.

\section{Conclusions}
\label{sec:concl}
We addressed the face clustering problem in the challenging domain of photo-streams acquired by a wearable camera. The problem at hand is complex to solve as we rely solely on face attributes in an image set captured under free-living conditions. The proposed model, through employing a deep-matching technique grasps robust representation of the face similarities. Moreover by applying two inner-track and inter-track constraints, achieves a relatively high performance while outperforming the state-of-the-art methods. A new challenging dataset with bounding boxes annotations and subject identity labels has been released \footnote{https://www.dropbox.com/s/zwepzpqz9st75x5/ICIP-dataset.zip?dl=0}. 

%

\bibliographystyle{IEEEbib}
\bibliography{bibliography}

\end{document}